\documentclass{article}

\usepackage[final, nonatbib]{nips_2017}

\usepackage[utf8]{inputenc} 
\usepackage[T1]{fontenc}    
\usepackage{hyperref}       
\usepackage{url}            
\usepackage{booktabs}       
\usepackage{amsfonts}       
\usepackage{nicefrac}       
\usepackage{microtype}      
\usepackage{amsmath}
\usepackage{tikz}
\usepackage{pgfplots}
\usepackage{siunitx}
\usepackage[justification=centering]{caption}
\usepackage{subcaption}

\title{Quantifying the Effects of Enforcing Disentanglement on Variational Autoencoders}

\author{
  Momchil Peychev, Petar Veli\v{c}kovi\'{c}, Pietro Li\`{o} \\
  Department of Computer Science and Technology\\
  University of Cambridge\\
  \texttt{mpeychev@cantab.net, \{petar.velickovic, pietro.lio\}@cst.cam.ac.uk}
}

\begin{document}

\bibliographystyle{plain}

\maketitle

\begin{abstract}
	The notion of disentangled autoencoders was proposed as an extension to the variational autoencoder by introducing a disentanglement parameter $\beta$, controlling the learning pressure put on the possible underlying latent representations. For certain values of $\beta$ this kind of autoencoders is capable of encoding independent input generative factors in separate elements of the code, leading to a more interpretable and predictable model behaviour. In this paper we quantify the effects of the parameter $\beta$ on the model performance and disentanglement. After training multiple models with the same value of $\beta$, we establish the existence of consistent variance in one of the disentanglement measures, proposed in literature. The negative consequences of the disentanglement to the autoencoder's discriminative ability are also asserted while varying the amount of examples available during training.
\end{abstract}

\section{Introduction}

The exponential growth in data availability and the rapid increase of computational power in the past decade have allowed neural network based algorithms to achieve impressive practical results in the fields of computer vision \cite{NIPS2012_4824, DBLP:journals/corr/SchroffKP15}, natural language processing and generation \cite{DBLP:journals/corr/abs-1303-5778, DBLP:journals/corr/OordDZSVGKSK16}, and game playing \cite{AlphaGo} to mention a few, surpassing human performance on several complex tasks \cite{He:2015:DDR:2919332.2919814, AlphaGo}. Despite the undeniable potential of the deep learning approach, however, more research is required to better understand its limits \cite{DBLP:journals/corr/SzegedyZSBEGF13}.

This work primarily concerns the model of a disentangled autoencoder which represents a recent development towards building more transparent and interpretable generative models. It is capable of learning independent generating factors separately in the network, thus being more predictable in its behaviour. Given certain input data we might know what values to expect for the code and, conversely, small disturbances of the code result in expected changes of the output. We study the properties of this model with respect to changing the values of the disentanglement parameter $\beta$, measuring both its disentanglement level and discriminative ability.

Autoencoders have been part of the neural network field since the late 80s \cite{Bourlard1988, NIPS1993_798}. Because of their capability to perform dimensionality reduction, they are sometimes considered to do a more powerful non-linear Principal Component Analysis (PCA) \cite{Bishop:2006:PRM:1162264, Goodfellow-et-al-2016}. The disentangled autoencoder was initially introduced by Higgins \emph{et al.} \cite{DBLP:journals/corr/HigginsMGPUBML16, beta-VAE} and has ever since been applied in semi-supervised learning environments as well \cite{VAE-semi-supervised}. It can be considered a generalisation of the variational autoencoder devised by Kingma and Welling \cite{DBLP:journals/corr/KingmaW13}. Earlier attempts at disentangled factor learning were reported to either require a priori knowledge about the data generating factors \cite{DBLP:journals/corr/CheungLBO14, Hinton:2011:TA:2029556.2029562, Reed:2014:LDF:3044805.3045052}, or do not scale well \cite{DBLP:journals/corr/CohenW14a, 2012arXiv1210.5474D}.

\section{Background}

Kingma and Welling \cite{DBLP:journals/corr/KingmaW13} derived the variational autoencoder framework by rearranging the evidence lower bound (ELBO) so that the intractable posterior $p_\theta(\mathbf{z}|\mathbf{x})$ can be eliminated
\begin{gather}
\mathrm{ELBO}(\theta, \phi) = \log p_\theta(\mathbf{x}) - D_{KL}(q_\phi(\mathbf{z}|\mathbf{x}) \| p_\theta(\mathbf{z}|\mathbf{x})) \\
\mathrm{ELBO}(\theta, \phi) =  \mathbb{E}_{\mathbf{z} \sim q_\phi(\mathbf{z} | \mathbf{x})} [\log p_\theta(\mathbf{x} | \mathbf{z})] - D_{KL}(q_\phi(\mathbf{z}|\mathbf{x}) \| p_\theta(\mathbf{z}))
\label{eqn:elbo-vae}
\end{gather}

The first term corresponds to the autoencoder's reconstruction error and the second one is the Kullback-Leibler (KL) divergence between the posterior approximation $q_\phi(\mathbf{z}|\mathbf{x})$ and $p_\theta(\mathbf{z})$, acting as a regulariser. In practice, this cost is typically dominated by the reconstruction error so Higgins \emph{et al.} \cite{DBLP:journals/corr/HigginsMGPUBML16, beta-VAE} took this approach further by specifying the optimisation problem
\begin{equation}
(\phi, \theta) = \max\limits_{\phi, \theta}
\mathbb{E}_{\mathbf{z} \sim q_{\phi}(\mathbf{z}|\mathbf{x})}[\log p_{\theta}(\mathbf{x}|\mathbf{z})] \text{\hspace{0.25cm}subject to\hspace{0.25cm}} D_{KL}(q_{\phi}(\textbf{z}|\textbf{x})||p_\theta(\textbf{z})) < \epsilon
\label{eqn:cond}
\end{equation}
for $\epsilon > 0$. Applying the Karush-Kuhn-Tucker conditions \cite{Bishop:2006:PRM:1162264}, Equation (\ref{eqn:cond}) can be written as a Lagrangian
\begin{equation}
\mathcal{L}(\theta, \phi; \textbf{x}) =
\mathbb{E}_{\mathbf{z} \sim q_{\phi}(\textbf{z}|\textbf{x})}[\log p_{\theta}(\textbf{x}|\textbf{z})] - \text{$\beta$\hspace{0.5mm}} D_{KL}(q_{\phi}(\textbf{z}|\textbf{x})||p_\theta(\textbf{z}))
\end{equation}
with $\beta \geq 0$, deriving the final disentangled autoencoder cost function. A practical choice is to set $q_{\phi}(\textbf{z}|\textbf{x}) = \mathcal{N}(\boldsymbol{\mu}, \boldsymbol{\sigma}^2\mathbf{I})$ and $p_\theta(\mathbf{z}) = \mathcal{N}(\textbf{0}, \mathbf{I})$. In this way not only the $D_{KL}$ term can be evaluated analytically \cite{DBLP:journals/corr/KingmaW13}, but choosing $p_\theta(\mathbf{z})$ to be the isotropic normal distribution with perfectly uncorrelated components forces the model to learn representations which encode statistically independent features about the data separately, in different positions of the code. Varying the value for $\beta$ regulates the amount of the applied learning pressure and in the next section we closely examine the effect of varying this disentanglement parameter.

\section{Experiments}

\subsection{Disentanglement level with respect to $\beta$}

\subsubsection{Data}

Higgins \emph{et al.} \cite{DBLP:journals/corr/HigginsMGPUBML16} made the key assumption that the observed data should possess transform continuities in order to be able to find some regularity in it in an unsupervised manner. We assume the input data is generated by factors of variation, densely sampled from their respective continuous distributions. In accordance to this considerations, we have constructed a synthetic dataset of 64x64 binarised images containing each a single shape. The generative factors defining each image are: a shape -- square ($\square$), ellipse (\tikz \draw (0,0) ellipse (6pt and 3pt);) or triangle ($\triangle$); position X (16 values); position Y (16 values); scale (6 levels); rotation (60 values over the $[0, \pi]$ range). The images were randomly separated in training, validation and test sets in a ratio 70:15:15 in a stratified way. Special care was taken to reduce the leakage between the subsets by removing duplicate images incidentally caused by some idempotent transformations (e.g. rotation of a square in \ang{90}, \ang{180} or \ang{270} produces the same figure). The final dataset consists of 267,021 images\footnote{The source code to reproduce all of our experiments described in this work can be found at \\ \url{www.github.com/mpeychev/disentangled-autoencoders}.}.

\subsubsection{Measuring disentanglement}

The disentanglement of an autoencoder cannot be usefully measured by its reconstruction accuracy or the KL-divergence term of the loss function as they fail to convey the notion of independence we want to obtain for the elements of the code. Precisely, disentanglement effect would mean distinguishing the generating factors of the data and encoding them in separate code elements.

Higgins \emph{et al.} \cite{DBLP:journals/corr/HigginsMGPUBML16} proposed a disentanglement measuring method which tries to evaluate this property of the trained autoencoders. A random set of generating factors is taken, the image $img_1$ is constructed, and the code means $\mathbf{z}^\mu_1 = encoder(img_1)$ are extracted. The same procedure is repeated, but this time one of the factors is randomly modified while all the others are kept the same. Denote the newly extracted code means with $\mathbf{z}^\mu_2$. A low capacity linear classifier is trained to map $\frac{|\mathbf{z}_1^\mu - \mathbf{z}_2^\mu|}{\max(|\mathbf{z}_1^\mu - \mathbf{z}_2^\mu|)}$ (division intended for normalisation) to the single factor that was changed during the process of obtaining $\mathbf{z}_1^\mu$ and $\mathbf{z}_2^\mu$. The classifier accuracy is then reported as a disentanglement measure of the autoencoder of interest. The assumption is that if a simple classifier is capable of inferring the single input generating factor responsible for the code perturbation, then the model provides some form of transparency and interpretability.

An alternative method in which one of the factors is fixed while all the others are randomly sampled between the generation of $img_1$ and $img_2$ is presented in a subsequent work by the same authors \cite{beta-VAE} but we only evaluate the first approach here.

\subsubsection{Results}

\begin{figure}[h]
	\centering
	\begin{subfigure}[t]{.45\linewidth}
		\centering
		\includegraphics[width=1\linewidth]{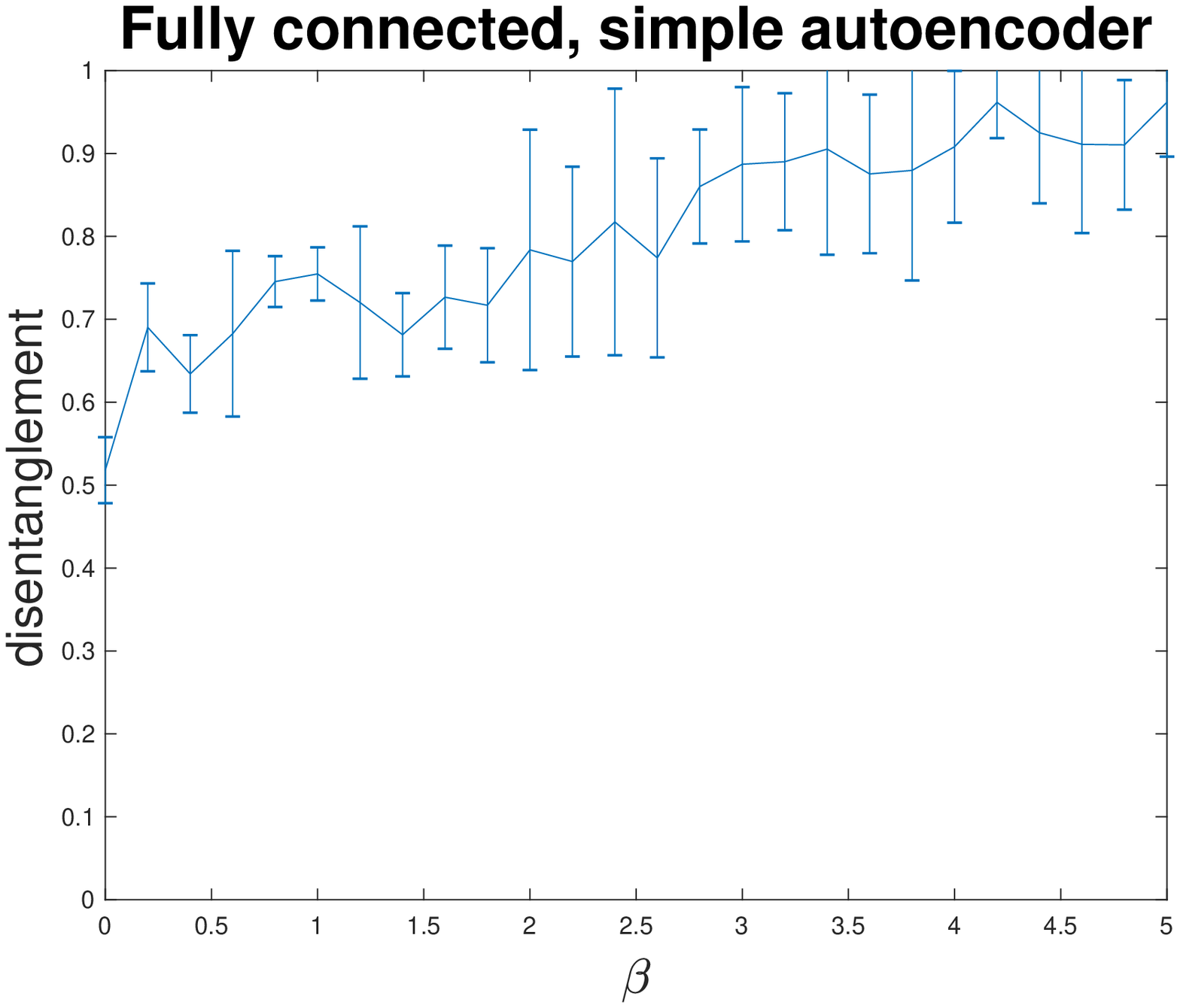}
	\end{subfigure}
	\begin{subfigure}[t]{.45\linewidth}
		\centering
		\includegraphics[width=1\linewidth]{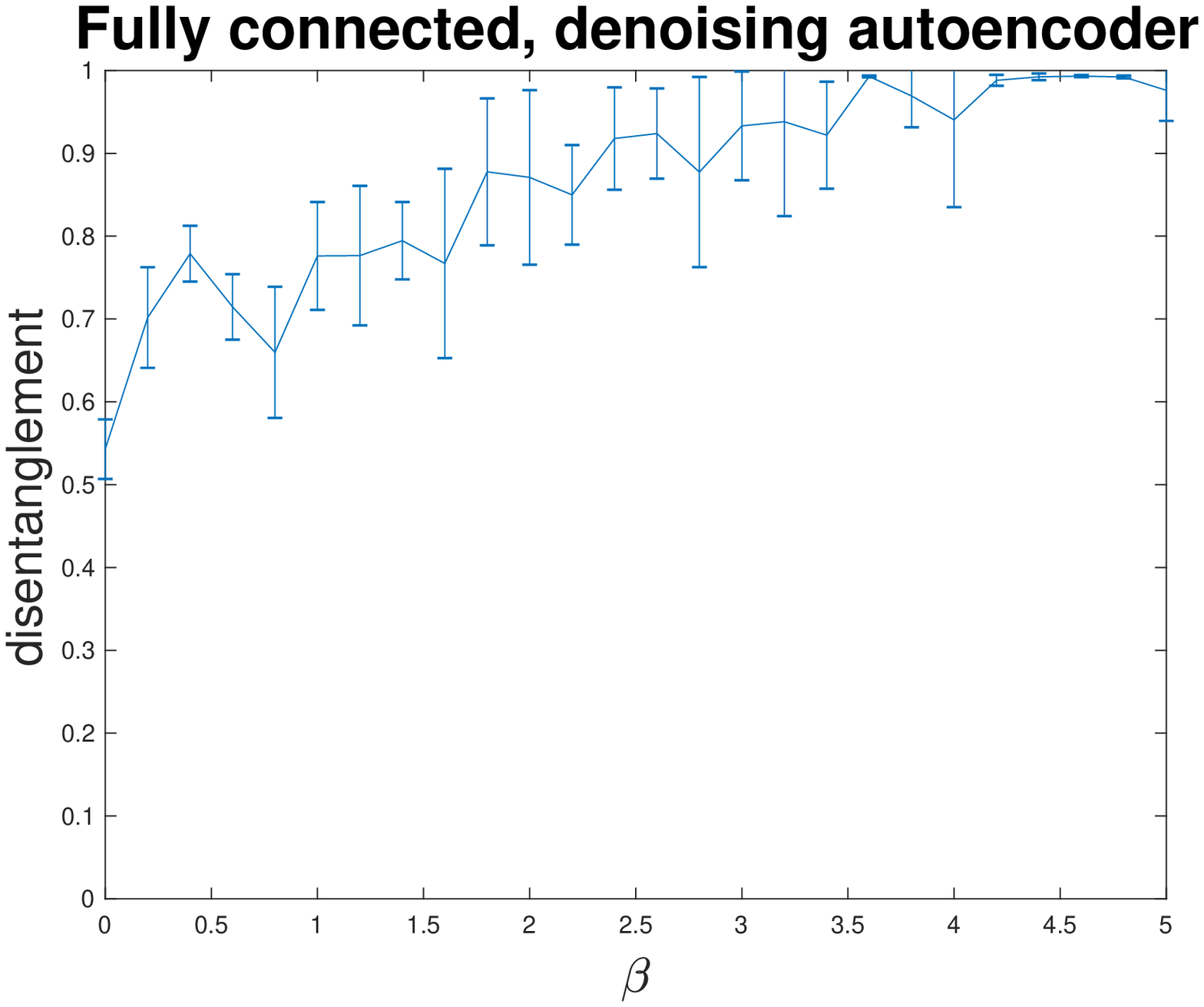}
	\end{subfigure}
	\vskip\baselineskip
	\begin{subfigure}[t]{.45\linewidth}
		\centering
		\includegraphics[width=1\linewidth]{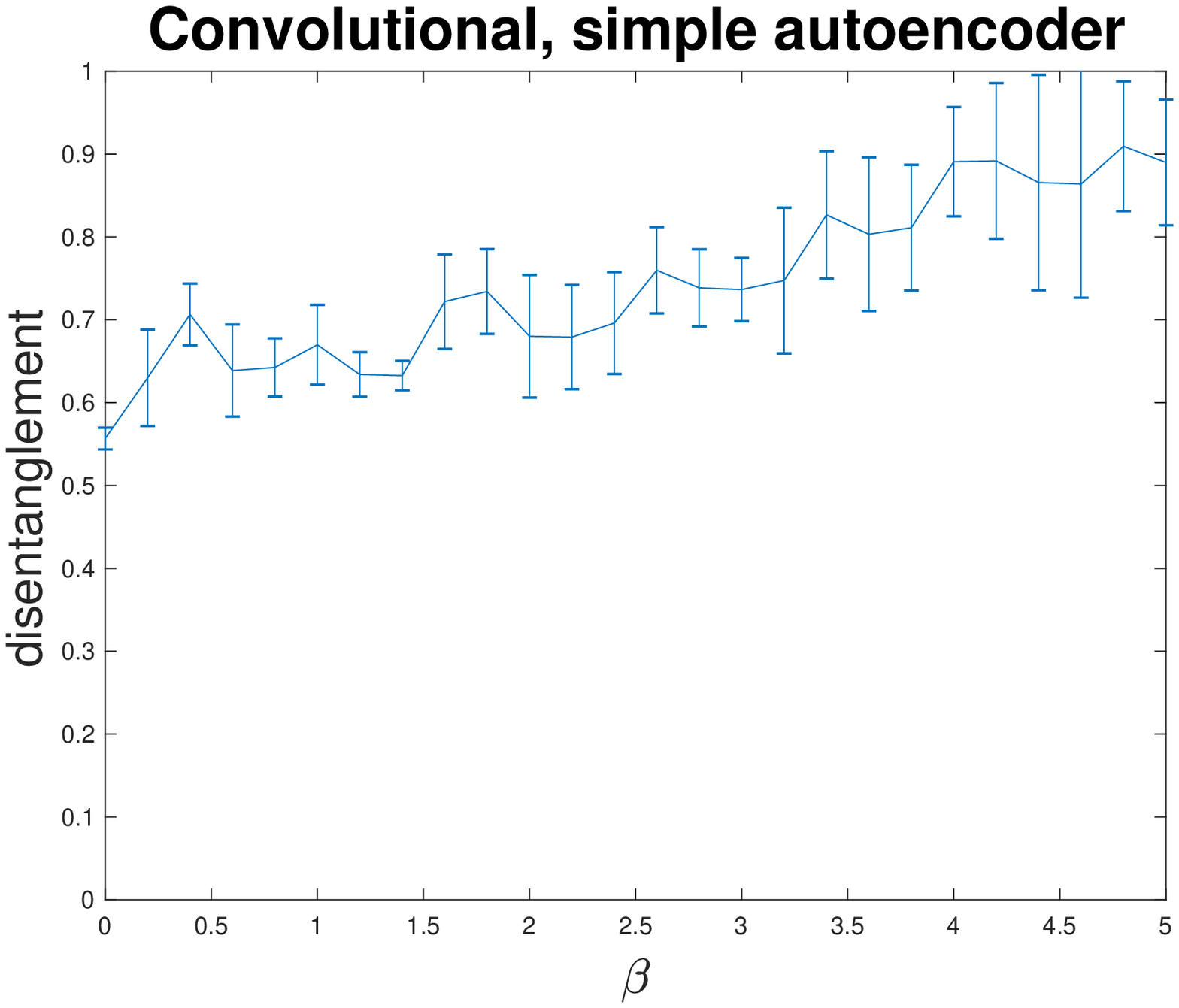}
	\end{subfigure}
	\begin{subfigure}[t]{.45\linewidth}
		\centering
		\includegraphics[width=1\linewidth]{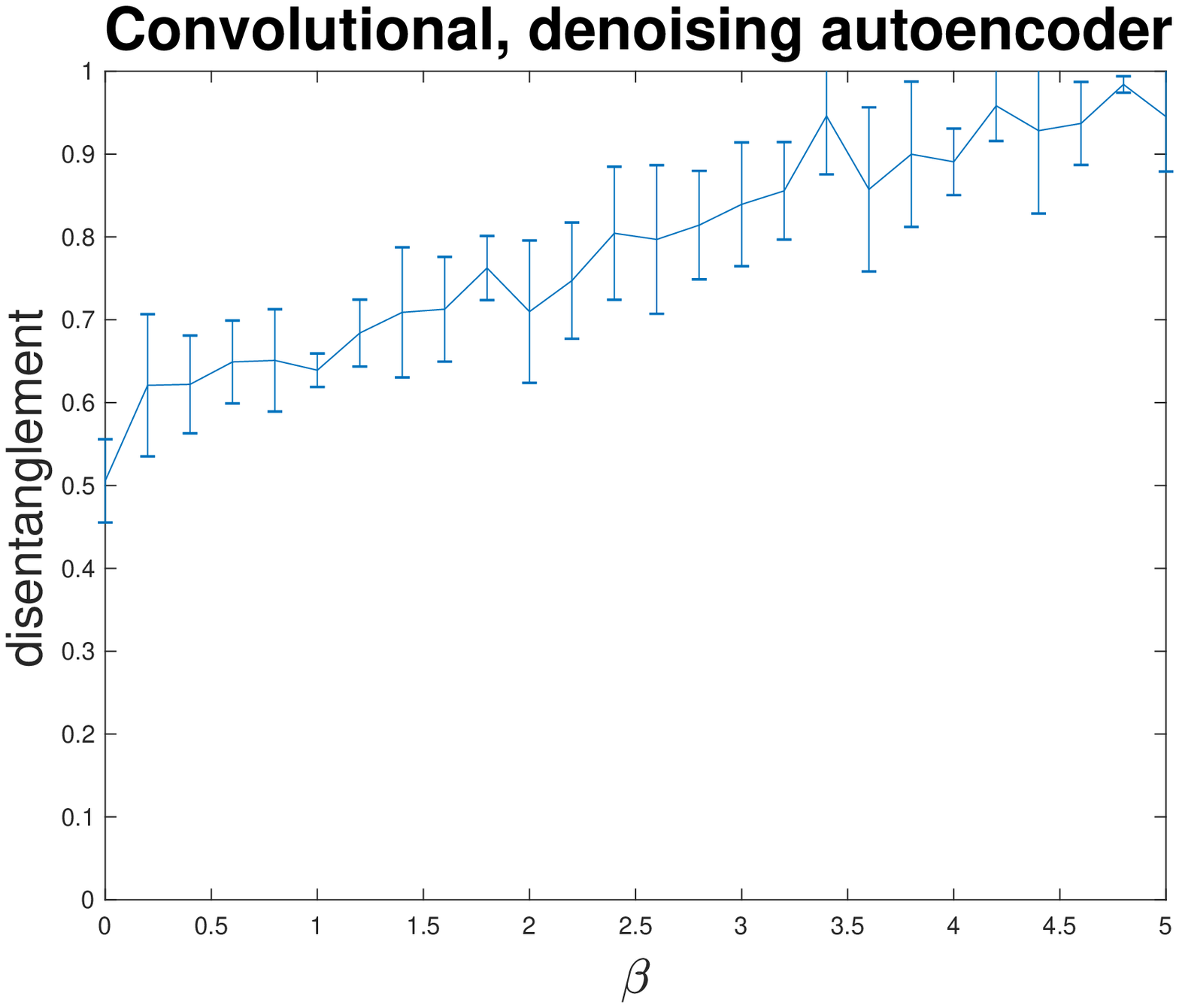}
	\end{subfigure}
	
	\caption{Disentanglement levels of the autoencoders, trained on the synthetic dataset, with respect to the parameter $\beta$.}
	\label{fig:disentanglement-level}
\end{figure}

The disentanglement levels of four types of autoencoders we trained, varying $\beta$ from 0 to 5 with a step of $0.2$ are presented in Figure \ref{fig:disentanglement-level}. Simple and denoising variants of autoencoders were considered and both fully connected and convolutional architectures were tested. The applied noise in the denoising case was salt-and-pepper, randomly flipping up to 20\% of the pixels. For each $\beta$, 5 autoencoder models were trained. In all graphs here and below we plot the means of the results obtained for all models trained with the same $\beta$ while the error bars denote standard deviations.

The first thing to become clear is the high variance between separate runs with the same value for $\beta$. A potential reason for that might be the method not being completely capable of closing the gap between the notion of disentanglement and factor independence we have with the underlying properties of the representations learnt by the autoencoders. For example, it was observed that for the position latents in the case of $\beta=4$ (which is supposed to be the disentangled case according to Higgins \emph{et al.} \cite{DBLP:journals/corr/HigginsMGPUBML16}), the autoencoder may sometimes learn ``curved'' or rotated, but still orthogonal, coordinate systems, which differs from what we would expect. Moreover, when reporting their results in \cite{DBLP:journals/corr/HigginsMGPUBML16}, the bottom 50\% of the obtained measurements have been discarded for unknown reasons (this was not performed when organising our results in Figure \ref{fig:disentanglement-level}). Higgins \emph{et al.} \cite{DBLP:journals/corr/HigginsMGPUBML16, beta-VAE} report results for fixed values of $\beta$ ($\beta=0, 1, $ and $4$) only, so to the best of our knowledge the findings about the intermediate values, presented in Figure \ref{fig:disentanglement-level} constitute previously unpublished work.

Another trend is the increase in the disentanglement with bigger values for $\beta$. The growth seems to be the most steady for convolutional denoising autoencoders. This is consistent with the claims that convolutional networks might be better at capturing image structures than fully connected ones and that adding noise and reconstructing the original data could act as a good regulariser.

The assumption for bigger values of $\beta$ is that at some point the autoencoder disentanglement will get flat (as starting to happen for the fully connected denoising case) and from then onwards further increase of $\beta$ will be damaging, as it will come at the cost of reducing the autoencoder's reconstruction ability. This in turn can lead to losing some useful learnt properties about the data. An application in which even small (but nonzero) values of $\beta$ can be harmful is described in the next section.

\subsection{MNIST classification with disentangled autoencoders}
\begin{figure}[h]
	\centering
	\begin{subfigure}[t]{.45\linewidth}
		\centering
		\includegraphics[width=1\linewidth]{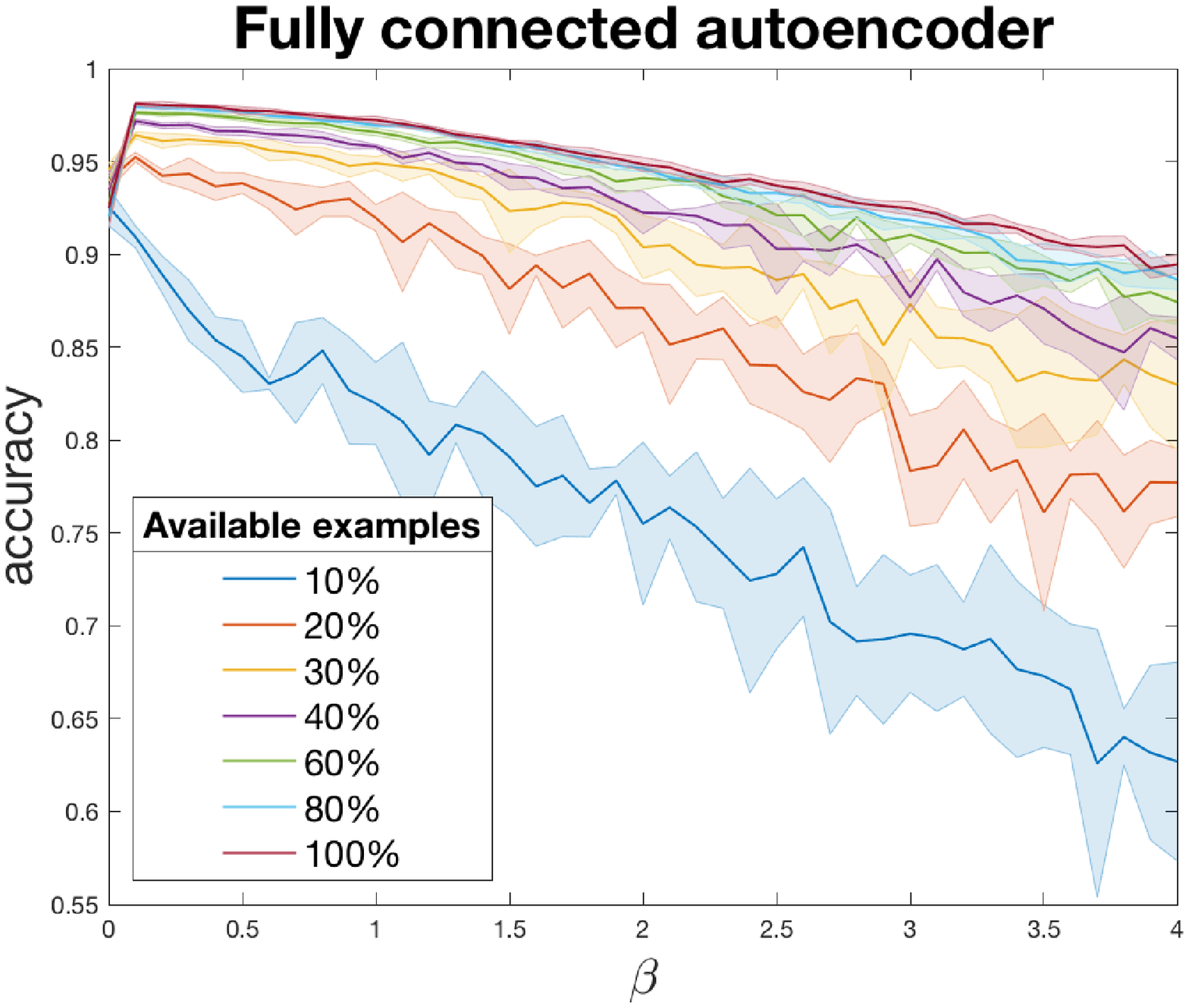}
	\end{subfigure}
	\begin{subfigure}[t]{.45\linewidth}
		\centering
		\includegraphics[width=1\linewidth]{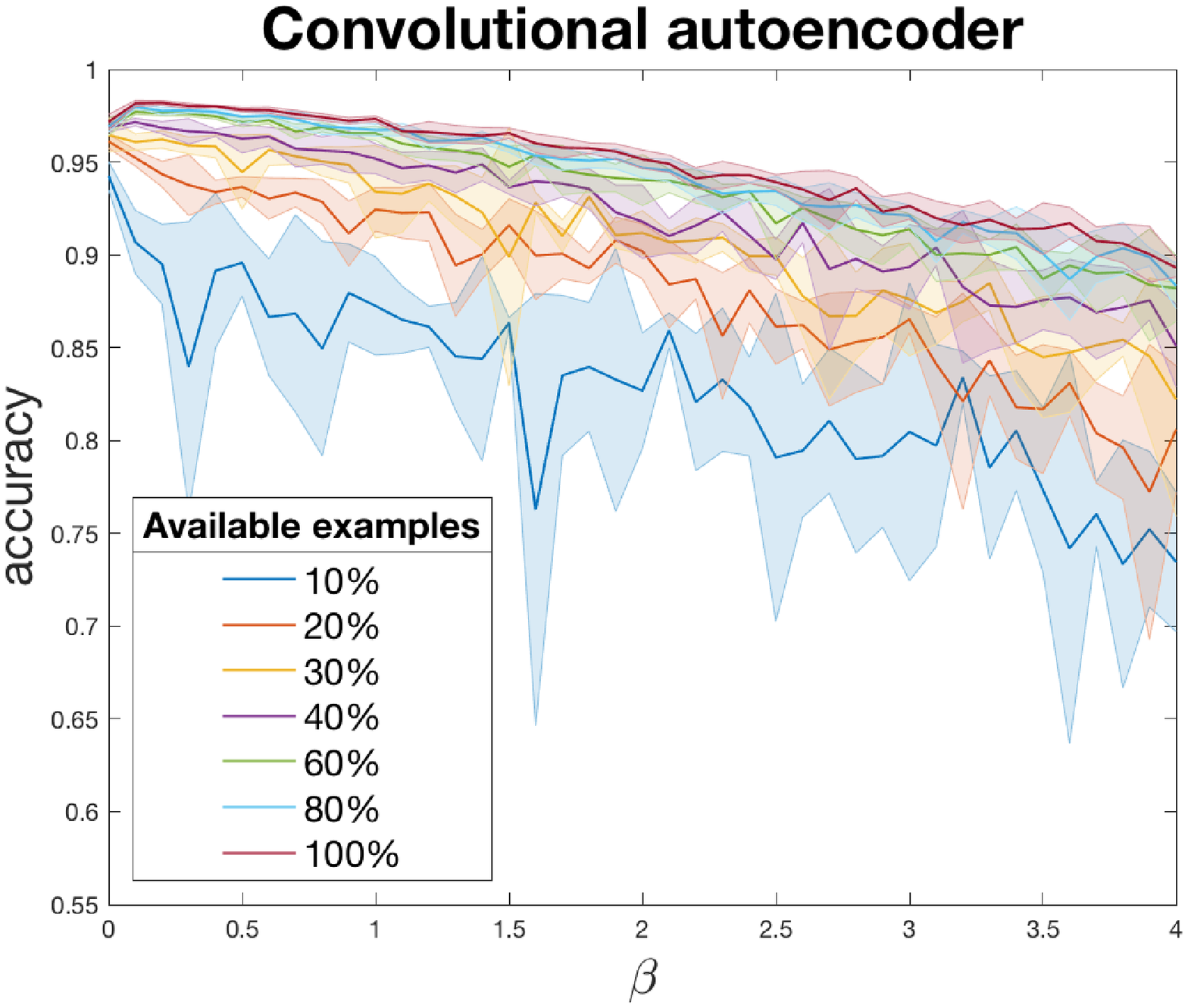}
	\end{subfigure}
	
	\caption{Results of MNIST classification with autoencoders. $\beta$ goes from 0 to 4 with step $0.1$. We execute the same experiments with different number of labels available during the training.\protect\footnotemark}
	\label{fig:mnist-classifiation}
\end{figure}

After evaluating disentangled autoencoders' behaviour on our synthetic dataset, it was a natural continuation to test them against an established machine learning benchmark. As such, the MNIST \cite{726791} dataset was considered to be a suitable candidate. The evaluation procedure began with an unsupervised autoencoder training first. Subsequently, a Support Vector Machine classifier was trained (using the same training dataset) to map the image codes, produced by the encoder network, to the respective image classes. The results are presented in Figure \ref{fig:mnist-classifiation}. The outcomes of the experiments using 10 and 30\% of the MNIST training dataset were included because the 20\% ones were outlying.

Increasing the number of training examples consistently increases the classification accuracy, as expected -- providing more labels helps the model generalise. Although with higher variance, the convolutional architecture seem to be more robust to training the autoencoder with fewer datapoints. The major spikes in the classification accuracy from $\beta=0$ to $\beta=0.1$ can be attributed to overfitting. When $\beta = 0$, the network tends to learn a one-to-one mapping and the latents may end up unrelated to their nearby values. The KL term acts as a regulariser, adding smoothness to the learnt latent manifold.

Taking into account the increasing error in the autoencoders reconstruction precision, it can be concluded that the autoencoder disentanglement is deteriorating for classification problems when applied to the MNIST dataset. This is an expected result, especially because of the lack of explicit continuity and generating factors of the MNIST images. However, it establishes the fact that there is a trade-off between the two terms of the disentangled autoencoder loss function and that they force the model to learn different properties about the data. When training a disentangled autoencoder, this trade-off should be considered and a balanced solution is desirable.

\footnotetext{Best viewed in colour.}

\section{Conclusion}

This work contributes with, to the best of our knowledge, new and unpublished findings about the properties of disentangled autoencoders. In particular, their level of disentanglement was measured over a whole range of values $\beta$ and it was discovered that, as expected, the disentanglement typically makes the models' performance worse in classification tasks.

\end{document}